\documentclass{article}
\usepackage{ijcai23}

\usepackage{times}
\usepackage{soul}
\usepackage{url}
\usepackage[hidelinks]{hyperref}
\usepackage[utf8]{inputenc}
\usepackage[small]{caption}
\usepackage{graphicx}
\usepackage{amsfonts}
\usepackage{amsmath}
\usepackage{amsthm}
\usepackage{booktabs}
\usepackage{algorithm}
\usepackage{algpseudocode}
\usepackage[switch]{lineno}
\usepackage{bm}
\usepackage{xcolor}

\title{Preserving Fairness in AI under Domain Shift}

\author{
    Serban Stan and Mohammad Rostami
    \affiliations
    University of Southern California
    \emails
    \{rostamim,sstan\}@usc.edu
}

\begin{document}

\maketitle

\begin{abstract}
    Existing algorithms for ensuring fairness in AI use a single-shot training strategy, where an AI model is trained on an annotated training dataset with sensitive attributes and then fielded for utilization. This training strategy is effective in problems with stationary distributions, where both training and testing data are drawn from the same distribution. However, it is vulnerable with respect to distributional shifts in the input space that may occur after the initial training phase. As a result, the time-dependent nature of data can introduce biases into the model predictions. Model retraining from scratch using a new annotated dataset is a naive solution that is expensive and time-consuming. We develop an algorithm to adapt a fair model to remain fair under domain shift using solely new unannotated data points. We recast this learning setting as an unsupervised domain adaptation problem. Our algorithm is based on updating the model such that the internal representation of data remains unbiased despite distributional shifts in the input space. We provide extensive empirical validation on three widely employed fairness datasets to demonstrate the effectiveness of our algorithm.
\end{abstract}

\section{Introduction} 

AI is being used extensively for technical tasks such as navigation in autonomous driving or recommender systems for online shopping. The successful deployment of AI in these real-world domains and applications has inspired the adoption of AI for many societal-related decision-making processes. Such tasks include processing loan applications, parole decisions, healthcare, and 
police deployments~\cite{chouldechova2018frontiers}. The recent success of AI stems from the re-emergence of deep learning, allowing for the optimization of increasingly complex decision functions in the presence of large datasets. A significant benefit of this data-driven learning approach is that it relaxes the need for tedious feature engineering. However, deep learning methods also have their drawbacks, e.g., data annotation is a time-consuming and expensive process~\cite{rostami2018crowdsourcing}.

Data-driven learning can also lead to 
training unfair models due to inherent biases existing in collected training datasets or skewed data distributions conditioned on certain features ~\cite{buolamwini2018gender}. Training models by simply minimizing the empirical error on relevant datasets may add spurious correlations between majority subgroup features and positive outcomes because statistical learning primarily discovers correlations rather than causation. Thus, the decision boundary of AI models may be informed by group-specific characteristics independent of the decision task ~\cite{Dua:2019}, which produces adverse outcomes for disadvantaged groups. For example, the income level is positively correlated with the male gender, which results in bias for unfair decisions against female applicants.

The above crucial concern on fairness for AI has resulted in significant research interest from the AI community. The first attempt to address bias in AI is to arrive at a commonly agreed-upon definition of fairness. Pioneer works in this area focused on defining quantitative notions for fairness based on commonsense intuition and using them to empirically demonstrate the presence and severity of bias in AI ~\cite{buolamwini2018gender,caliskan2017semantics}. Most existing fairness metrics consider the input data points assuming characteristics of protected subgroups~\cite{feldman2015certifying}, e.g., gender and race, in addition to standard features used for classification. Based on subgroup membership, majority and minority populations emerge, which should be treated to equal outcomes under fair learning. A model is then assumed to be fair if its predictions possess a notion of conditional probabilistic independence for data membership into the subgroups~\cite{mehrabi2021survey} (see the \textit{Fairness Metrics} section for definitions of common fairness metrics). 

A fair model can be trained by minimizing the model bias using these metrics, e.g., demographic parity, in addition to minimizing the empirical risk on a given training dataset. Despite effectively mitigating bias, most existing fair model training algorithms consider that the data distribution will remain stationary after the training stage. However, this assumption is rarely true in practical settings, particularly when a model is used over extended periods~\cite{rostami2020using} in dynamic societal applications. As a result, a fair model might fail to maintain fairness under the input-space distributional shifts or when the model is used on differently sourced datasets~\cite{pooch2019can,rostami2021lifelong}. The naive solution of retraining the model after distributional shifts requires annotating new data points to build datasets representative of the new input distribution. This process, however, is time consuming and expensive for deep learning. It is highly desirable to develop algorithms that can preserve model fairness under distribution shifts. Unfortunately, this problem has been marginally explored in the literature.

The problem of model adaptation under distributional shifts has been investigated extensively in the unsupervised domain adaptation (UDA) literature ~\cite{tzeng2017adversarial,daume2009frustratingly}. The goal of UDA is to train a classification model with a good generalization performance for a target domain where only unannotated data is available through transferring knowledge from a related source domain, where annotated data is accessible. A primary group of UDA algorithms achieves this goal by matching the source, and the target distributions in an embedding space ~\cite{10.1007/978-3-319-71246-8_45} such that the embedding space is domain-agnostic. As a result, a classifier that receives its input as data representations in the embedding space will generalize well in the target domain, despite being trained solely using the source domain annotated data. To align the two distributions in such an embedding, data points from both domains are mapped into a shared feature space that is modeled as the output space of a deep neural network encoder. The deep encoder is then trained to minimize the distance between the two distributions, measured in terms of a suitable probability distribution metric. However, existing UDA algorithms overlook model fairness and solely consider improving model performance in the target domain. In this work,  we adopt the idea of domain alignment to preserve model fairness and mitigate model biases introduced by domain shift.

\textbf{Contribution:} We address the problem of preserving model fairness under distributional shifts in the input space when only unannotated data is accessible. We model this problem within the domain adaptation paradigm. We consider the initial annotated training data as the source domain and the collected unannotated data during model execution as the target domain. Our contribution is to develop an algorithm that minimizes distributional mismatches in a shared embedding space, modeled as the output space of a deep encoder. We present empirical results using tasks built from three standard fairness benchmark datasets to demonstrate the applicability of the proposed algorithm.

\section{Related Work}
Our work is related to works in   UDA and fairness in AI.

\subsection{Fairness in AI}

There are various approaches to training a fair model for a single domain. A primary idea in existing works is to map data points into an embedding space at which the sensitive attributes are entirely removed from the representative features, i.e., an attribute-agnostic space for which fairness is measured at the classifier output using a desired fairness metric. As a result, a classifier that receives its input from this space will make unbiased decisions due to the independence of its decisions from the sensitive attributes. Ray et al.~\cite{jiang2020wasserstein} develop a fairness algorithm that induces probabilistic independence between the sensitive attributes and the classifier outputs by minimizing the optimal transport distance between the probability distributions conditioned on the sensitive attributes. The transformed probability in the embedding space then becomes independent (unconditioned) from the sensitive attributes. 
Celis et al.~\cite{celis2019classification} study the possibility of using a meta-algorithm for fairness with respect to several disjoint sensitive attributes.
 Du et al.~\cite{du2021fairness} have followed a different approach. Instead of training an encoder that removes the sensitive attributes in a latent embedding space and then training a classifier, they propose to debias the classifiers by leveraging samples with the same ground-truth label yet having different sensitive attributes. The idea is to discourage undesirable correlation between the sensitive attribute and predictions in an end-to-end scheme, allowing for the emergence of attribute-agnostic representations in the hidden layers of the model. 
 
 Beutel et al.~\cite{beutel2017data} benefit from removing sensitive attributes to train fair models by indirectly enforcing decision independence from the sensitive attributes in a latent representation using adversarial learning. They also amend the encoder model with a decoder to form an autoencoder. Since the representations are learned such that they can self-reconstruct the input, they become discriminative for classification purposes as well. Our work builds upon using adversarial learning to preserve fairness when distribution shifts exist. In order to combat domain shift, our idea is to additionally match the target data distribution with the source data distribution in the latent embedding space, a process that ensures classifier generalization.

\subsection{Unsupervised Domain Adaptation}
Works on domain alignment for UDA follow a diverse set of strategies. The goal of existing works in UDA is solely to improve the prediction accuracy in the target domain in the presence of domain shift without exploring the problem of fairness. The closest line of research to our work addresses domain shift by minimizing a probability discrepancy measure between two distributions in a shared embedding space. Selection of the discrepancy measure is a critical task for these works. Several UDA methods simply match the low-order empirical statistics of the source and the target distributions as a surrogate for the distributions. For example, the Maximum Mean Discrepancy (MMD) metric is defined to match the means of two distributions for UDA~\cite{long2015learning,long2017deep}. Correlation alignment is another approach to include second-order moments~\cite{sun2016deep}. Matching lower-order statistical moments overlooks the existence of discrepancies in higher-order statistical moments. In order to improve upon these methods, a suitable probability distance metric can be incorporated into UDA to consider higher-order statistics for domain alignment. A suitable metric for this purpose is the Wasserstein distance (WD)  ~\cite{courty2017optimal,damodaran2018deepjdot}. Since WD possesses non-vanishing gradients for two non-overlapping distributions, it is a more suitable choice for deep learning than more common distribution discrepancy measures, e.g., KL-divergence. Optimal transport can be minimized as an objective using first-order optimization algorithms for deep learning.
Using WD has led to a considerable performance boost in both single-source~\cite{damodaran2018deepjdot,stan2022domain} and multi-source UDA~\cite{stansecure}.

\subsection{Domain Adaptation in Fairness}
Despite being an important technique to address practically significant challenges, works on benefiting from knowledge transfer for domain adaptation in fairness are relatively limited. Madras et al.~\cite{madras2018learning} benefit from adversarial learning to learn domain-agnostic transferable representations for fair model generalization. Coston et al.~\cite{coston2019fair} consider a UDA setting where the sensitive attributes for data points are accessible only in one of the source or the target domains. Their idea is to use a weighted average to compute the empirical risk and then tune the corresponding data point-specific weights to minimize co-variate shifts. Schumann et al.~\cite{schumann2019transfer} consider a similar setting, define the fairness distance of equalized odds, and then use it as a regularization term in addition to empirical risk, minimized for fair cross-domain generalization. Hu et al.~\cite{hu2019distributed} address fairness in a distributed learning setting, where the data exist in various servers with private demographic information. Singh et al.~\cite{singh2021fairness} consider that a causal graph for the source domain data and anticipated shifts are given. They then use feature selection to estimate the fairness metric in the target domain for model adaptation. Zhang and Long~\cite{zhang2021assessing}  explore the possibility of training fair models in the presence of missing data in a target domain using a source domain with complete data and find theoretical bounds for this purpose.
Our learning setting is relevant yet different from the above settings. We consider a standard UDA setting where the sensitive attributes are accessible in both domains. The challenge is to adapt the model to preserve fairness in the target domain.

\section{Problem Formulation}
We formulate our problem in a classic UDA setting.
Let $(X, A, Y) \in \mathbb{R}^d \times \{0, 1\} \times \{0, 1\}$ be the format of datasets. $X \in \mathbb{R}^n$ represents a feature vector with $n$ entries, corresponding to different dataset characteristics that are used for prediction, e.g., occupation length, education years, credit history etc. $A \in \{0, 1\}$ represents a sensitive attribute for which we aim to train a fair model, e.g., sex, race, age. Finally, $Y \in \{0,1\}$ represents the binary label of our dataset, e.g., approved for credit, income greater than $50k$ etc. 

Given a fully annotated dataset in the source domain $(X^s, A^s, Y^s)$, we can train a fair model $f_\theta : (X^s, A^s) \rightarrow Y^s$ with learnable parameter $\theta$. Let $P_{\mathcal{S}}(\bm{X},\bm{A})^{s}$ denotes the unknown input feature distribution in the source domain. In a normal learning setting, we search for the optimal parameter $\theta^*$ such that it only minimizes the true risk, i.e., 
 $\bm{\theta}^* =\arg\min_{\bm{\theta}}\{\mathbb{E}_{(\bm{x}^{s}, \bm{a}^s) \sim P_{\mathcal{S}}(\bm{X},\bm{A})^{s}}(\mathcal{L}(f_{\bm{\theta}}(\bm{x}^{s}, \bm{a}^s),\bm{y}^{s})\}$ for some suitable loss function $\mathcal{L}$, e.g., cross-entropy. Given that we do not have direct access to the data distribution, we use ERM as a surrogate objective function and learn an approximation $\hat \theta$ of $\theta^*$ using the training dataset:
\begin{equation}
\small
\begin{split}
    & \hat{\bm{\theta}} =\arg\min_{\bm{\theta}}\{\frac{1}{N} \sum_{i=1}^N\mathcal{L}_{bce}(f_{\bm{\theta}}(\bm{x}^{s}, \bm{a}^s),\bm{y}^{s})\}, 
    \end{split}
\label{eq:ce_loss}
\end{equation} 
 where $N$ represents the number of samples in the dataset, 
  $\mathcal{L}_{bce}$ is the binary cross-entropy loss, and $\hat y=f_{\bm{\theta}}(\bm{x}^{s}, \bm{a}^s)$ represents the probabilistic output of the model. Solving Eq.~\eqref{eq:ce_loss} does not lead to obtaining a fair model, as only prediction accuracy is optimized. Bias in the training dataset, e.g., over/under-representation of subgroups can lead to unfair models, even if the sensitive attribute is hidden.

\begin{figure*}[ht]
    \centering
    \includegraphics[width=.65\textwidth]{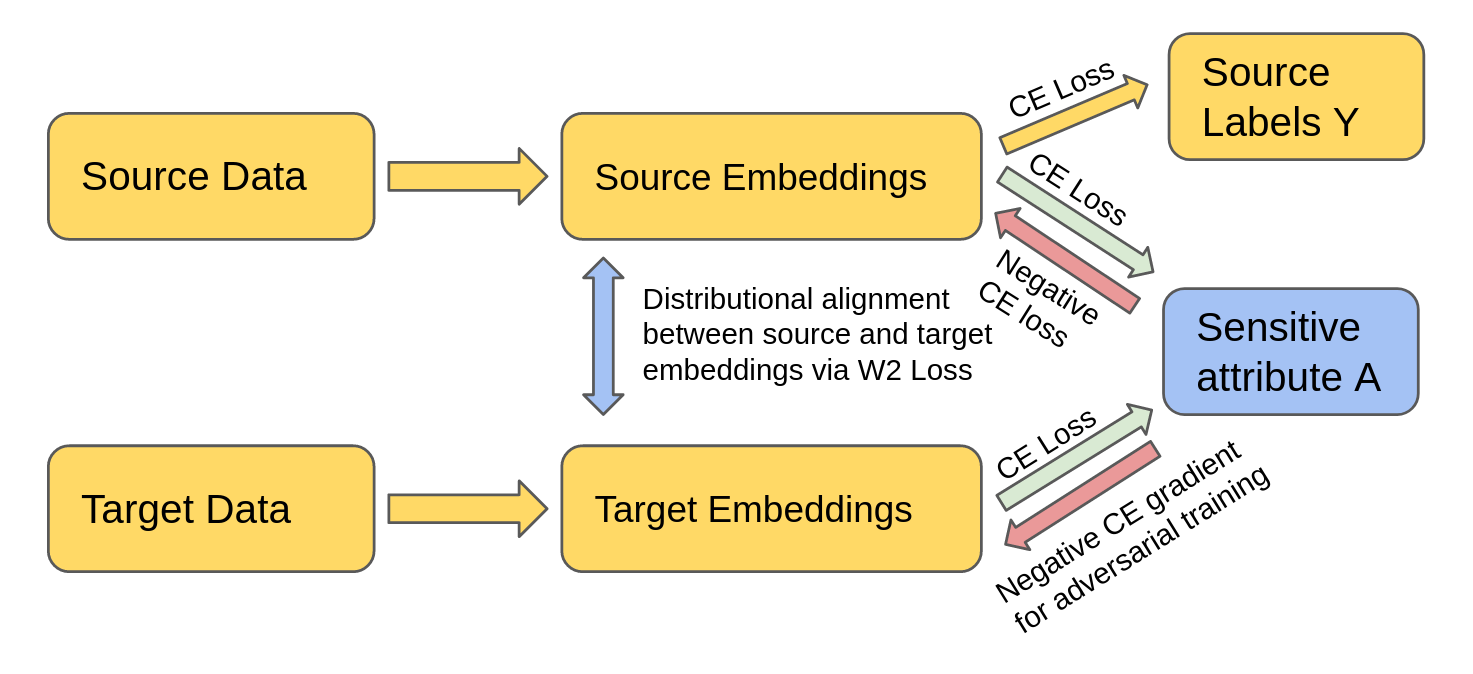}
    \caption{Block-diagram description of the proposed framework}
    \label{figure:method-description}
\end{figure*}

We consider that our classifier model $f_\theta : \mathbb{R}^d \rightarrow \mathbb{R}^2$ can be decomposed into an encoder subnetwork $e_u : \mathbb{R}^d \rightarrow \mathbb{R}^z$, with learnable parameters $u$,  followed by a classifier subnetwork $g_v : \mathbb{R}^z \rightarrow \mathbb{R}^2$ with learnable parameters $v$. In this formulation, $f_\theta(\cdot) = (g_v \circ e_u)(\cdot)$, $\theta = (u,v)$, $z$ denotes the size of the latent representation space that we want to make sensitive-agnostic. To induce fairness in the latent space, we consider an additional classification network $h_w : \mathbb{R}^z \rightarrow \mathbb{R}^2$ with learnable parameters $w$. This classifier is tasked to predict the sensitive attribute $A$ using the latent space features $e_u(\bm{x}^s, \bm{a}^s)$. The primary idea is to induce probabilistic independence from sensitive attributes by adversarially optimizing $g$ and $h$. Latent representations independent of the sensitive attribute, $A$, will lead to $h$ performing poorly, while fairness-agnostic representations will lead to $h$ correctly predicting the sensitive attribute for a specific input. 

Consider the loss for predicting sensitive attributes:
\begin{equation}
\begin{split}
    & \mathcal{L}_{fair} = \mathcal{L}_{bce}{((h_w \circ e_u)(\bm{x}^s, \bm{a}^s), \bm{a}^s)}.
\end{split}
\label{eq:ce_fair_loss}
\end{equation} 

From our discussion, we consider the following fairness-guaranteeing alternating minimization process: 

\begin{enumerate}
\item We fix the encoder $e_u$ and minimize the fairness loss $\mathcal{L}_{fair}$ through updating the attribute classifier $h_w$.
\item We then fix the attribute classifier $h_w$ and maximize the fairness loss $\mathcal{L}_{fair}$ by updating the encoder $e_u$.
\end{enumerate}
The first step will perform empirical risk minimization (ERM) for the fairness classifier, conditioned on the encoder. The second step will keep the classifier fixed, and ensure the latent data representations are as little informative as possible about the sensitive attribute $A$. Empirical exploration demonstrated that iterative alternation between these two optimization steps will force the encoder $e_u$ to produce latent representations that are independent from the sensitive attributes. This high-level idea is presented in Figure~\ref{figure:method-description}.

The above approach would suffice in practice if the target data is drawn from the source domain. In the case when target data is sampled from a domain $(X^t, A^t, Y^t)$ with $P_{\mathcal{T}}(\bm{X},\bm{A})^{t} \neq P_{\mathcal{S}}(\bm{X},\bm{A})^{s}$, the source trained model will perform poorly during testing due to domain shift. To improve generalization on the target domain, we can minimize the empirical distance between $e_u(\bm{x}^s, \bm{a}^s)$ and $e_u(\bm{x}^t, \bm{a}^t)$. Under this restriction, the source domain classifier $g_v$ will be able to generalize fairly on target domain. While this technique has been used in UDA, by itself it is not sufficient to guarantee fairness at adaptation time.

 
\section{Proposed Algorithm}

The block-diagram description of our proposed approach is presented in Figure \ref{figure:method-description}. Initially, we train a fair model on the source domain dataset $(X^s, A^s, Y^s)$ through the following iterative three-step procedure:

\begin{enumerate}
    \item We optimize the classifier  $(f_\theta = g_v \circ e_u)$ network with cross entropy loss in an end-to-end scheme by minimizing Eq. \ref{eq:ce_loss}. This process will generate latent features informative and discriminative  for   decision making.
    \item We then fix the feature extractor encoder subnetwork $e_u$ and optimize the sensitive attribute classifier $h_w$ by minimizing the loss in Eq. \ref{eq:ce_fair_loss}. This  step will enforce the sensitive attribute classifier to extract information from the representations in the embedding space that can be used for predicting the sensitive attribute $A$.
    \item We freeze the sensitive attribute classifier $h_w$ and update the encoder subnetwork $e_u$ in order to maximize the fairness loss function in Eq. \ref{eq:ce_fair_loss}. This step will force the encoder to produce   representations that are independent from the sensitive attribute $A$.
\end{enumerate}

To preserve fairness under distribution shift, the adaptation process relies on coupling the source and target domain through the shared embedding space between the two. If we match the distributions in the embedding space, the decision classifier $g_v$ will be able to generalize well in the target domain. Aligning the embedding distributions is sufficient for achieving this goal, i.e., $e(P_\mathcal{S}(\bm{X}, \bm{A})) \approx e(P_\mathcal{T}(\bm{X}, \bm{A}))$. 

We employ metric minimization from the UDA literature to enforce domain alignment \cite{Lee_2019_CVPR,10.1007/978-3-319-71246-8_45}. The idea is to select a suitable probability distribution distance $d(\cdot, \cdot)$ and minimize it for the encoder output, i.e. $d(e(P_\mathcal{S}(\bm{X}, \bm{A})) , e(P_\mathcal{T}(\bm{X}, \bm{A})))$, to guarantee a shared embedding feature space between source and target. The choice of the distribution distance $d(\cdot, \cdot)$ is an algorithmic design choice and various metric have been used for this purpose.
In this work, we use the Sliced Wasserstein Distance (SWD) \cite{10.1007/978-3-319-71246-8_45}. SWD is an approximation of the optimal transport metric and enhances the applicability of using stochastic-gradient based optimization~\cite{rostami2023}. Optimal transport only has a closed-form solution for $1D$ distributions. The idea behind the SWD approximation is to slice two high dimensional distributions to generate 1D distributions and then compute their distance as the average of these $1D$ WD slices, computed using several random projections to generate them. In addition to the closed form solution, we can compute SWD using the empirical samples from the two distributions. In the context of our setup, SWD can be computed as follows:
\begin{equation}
\small
\begin{split}
    &\mathcal{L}_{swd} = \frac{1}{K} \sum_{i=1}^{K} WD^1(\langle e(\bm{x}^s, \bm{a}^s), \gamma_i \rangle, \langle e(\bm{x}^t, \bm{a}^t), \gamma_i \rangle ),
\end{split}
\label{eq:swd_loss}
\end{equation} 
where, $WD^1(\cdot,\cdot)$ denotes the $1D$ WD distance, $K$ is the number of random projections we are averaging over and $\gamma_i$ is one such projection direction.

To implement domain alignment to preserve fairness in the target domain under distributional shifts, we augment steps $(1)-(3)$ described above with two additional steps:
\begin{enumerate}
    \setcounter{enumi}{3}
    \item We minimize the empirical SWD distance between $e(P_\mathcal{S}(\bm{X}, \bm{A}))$ and $e(P_\mathcal{T}(\bm{X}, \bm{A}))$ via Eq. \ref{eq:swd_loss}. This step ensures the source-trained classifier $g$ will generalize on samples from the target domain, i.e.,   $e(P_\mathcal{T}(\bm{X}, \bm{A}))$.
    \item We repeat steps (2) and (3) using solely the sensitive attributes of the target domain. 
\end{enumerate}
The above additional steps will update the model on the target domain to preserve both fairness and accuracy. Following steps (1)-(5), the loss function becomes:
\begin{equation}
\begin{split}
    &\mathcal{L}_{bce}(\hat y, y_{src}) + \alpha \mathcal{L}_{fair}^{src} + \beta \mathcal{L}_{fair}^{tar} + \gamma \mathcal{L}_{swd},
\end{split}
\label{eq:total_loss}
\end{equation} 
where the hyperparameter $\alpha, \beta,$ and $\gamma$ can be tuned using cross validation.
 We provide algorithmic pseudocode for our proposed fairness preservation process  in Algorithm \ref{algorithm:FairAdapt}. 

\begin{algorithm}[t]
\small
\caption{$\mathrm{FairAdapt}\left (\alpha , \beta, \gamma, thresh, ITR \right)$\label{algorithm:FairAdapt}} 
 {
    \begin{algorithmic}[1]
        \For {$itr = 1,\ldots, ITR$ }
            \State \textbf{Source Training}: 
            \State Optimize $\alpha \mathcal{L}_{bce}$ via Eq. \ref{eq:ce_loss}.
            \State Optimize $\beta \mathcal{L}_{fair}$ via Eq. \ref{eq:ce_fair_loss} and freezing $u$. 
            \State Optimize $-\beta \mathcal{L}_{fair}$ via Eq. \ref{eq:ce_fair_loss} and freezing $h$. 
            
            \If {$itr > thresh$}
                \State \textbf{Target Adaptation}: 
                \State Optimize $\gamma \mathcal{L}_{swd}$ via Eq. \ref{eq:swd_loss}.
                \State Optimize $\beta \mathcal{L}_{fair}$ via Eq. \ref{eq:ce_fair_loss} and freezing $u$. 
                \State Optimize $-\beta \mathcal{L}_{fair}$ via Eq. \ref{eq:ce_fair_loss} and freezing $h$.
            \EndIf
        \EndFor
        \State \Return $u,g$
    \end{algorithmic}}
\end{algorithm}

\section{Empirical Validation}

Our learning setting is under explored. For this reason, we adopt existing datasets and tailor them for our formulation.

\subsection{Datasets and Tasks}

Datasets relevant to fairness tasks pose a binary decision problem e.g. approval of a credit application, alongside relevant features e.g. employment history, credit history etc. and group related sensitive attributes e.g. sex, race, nationality, age. Based on sensitive group membership, data points can be part of privileged or unprivileged groups. 

We perform experiments on three datasets widely used by the AI fairness community. We will consider \textit{sex} as our sensitive attribute, which is recorded for all three datasets. 

The \textbf{UCI Adult dataset}\footnote{https://archive.ics.uci.edu/ml/datasets/Adult} is part of the UCI database \cite{Dua:2019} and consists of 1994 US Census data. The task associated with the dataset is predicting whether annual income exceeds 50k. After data cleaning, the dataset consists of more than $48,000$ entries. Possible sensitive attributes for this dataset include \textit{sex} and \textit{race}.

The \textbf{UCI German credit dataset} \footnote{https://archive.ics.uci.edu/ml/datasets/statlog+(German+credit+data)} contains financial information for $1000$ different people applying for credit. The predictive task involves categorizing individuals as good or bad credit risks. \textit{Sex} and \textit{age} are possible sensitive attributes for the German dataset.

The \textbf{COMPAS recidivism dataset} \footnote{https://github.com/propublica/COMPAS-analysis/} maintains information of over $5,000$ individuals' criminal records. Models trained on this dataset attempt to predict people's two year risk of recidivism. For the COMPAS dataset, \textit{sex} and \textit{race} may be used as choice of sensitive attributes.

\begin{table*}[!ht]
    \centering
    \scalebox{.9}{
        \setlength\tabcolsep{2pt}
        \begin{tabular}{|l|c|c|c|c|c|c|c|c|c|c|c|c|}
        \hline
        \multicolumn{1}{|c|}{Split} & \multicolumn{4}{c|}{Source} & \multicolumn{4}{c|}{Target} \\
        \cline{2-9}
        & Size &Y=0 &A=0$|$Y=0 &A=0$|$Y=1 &Size &Y=0  &A=0$|$Y=0 &A=0$|$Y=1 \\
        \hline
        A & 34120 & 0.76 & 0.39 & 0.15 & 14722 & 0.76 & 0.39 & 0.15 \\
        A1 & 12024 & 0.53 & 0.41 & 0.16 & 5393 & 0.91 & 0.49 & 0.18 \\
        A2 & 29466 & 0.66 & 0.34 & 0.14 & 2219 & 0.97 & 0.48 & 0.30 \\
        A3 & 11887 & 0.52 & 0.42 & 0.16 & 778 & 0.89 & 0.39 & 0.17 \\
        \hline
        \end{tabular}
    }
    \caption{Data split statistics. A corresponds to the Adult dataset. The row with no number i.e. A corresponds to random data splits. The numbered rows e.g. A1,A2,A3 correspond to statistics for specific splits. The columns represent the probabilities of specific outcomes for specific splits e.g. $P(Y = 0)$. Results when using \textit{sex} as sensitive attribute.}
    \label{table:data-split-stats}
\end{table*}

\subsubsection{Evaluation Protocol Motivation}

Historically, experiments on these datasets have considered random $70/30$ splits for training and test. While such data splits are useful in evaluating overfitting for fairness algorithms, features for training and test sets will be sampled from the same data distribution. This assumption is unlikely in practice when deploying a fair model to a different dataset. We consider natural data splits obtained from sub-sampling the three datasets along different criteria. We show that compared to random splits, where learning a model that guarantees fairness on the source domain is often times enough to have the model guarantee fairness on the target domain, experiencing domain discrepancy between the source and target can lead fair models trained on the source domain to produce unfair or degenerate solutions on the target domain. In short, these splits introduce domain gap between the testing and training splits.
 For detailed explanations about the splits for each dataset, please refer to the supplementary material. 
 
Next, for each of the three datasets we will generate source/target data splits where ignoring domain discrepancy between the source and target can negatively impact fairness transfer. Per dataset, we will produce three such splits. We characterize the label distributions and sensitive attribute conditional distributions for the Adult dataset in Table \ref{table:data-split-stats}. We provide similar analysis for the German and COMPAS datasets in the supplementary material.

\textbf{Adult Dataset}. We use age, education and race to generate source and target domains. This can be a natural occurrence in practice, as gathered census information may differ along these axes geographically. For example, urban population is on average more educate than rural population \footnote{https://www.ers.usda.gov/topics/rural-economy-population/employment-education/rural-education/}, and more ethnically diverse \footnote{https://www.ers.usda.gov/data-products/chart-gallery/gallery/chart-detail/?chartId=99538}. Thus, a fair model trained on one of the two populations will need to overcome distribution shift when evaluated on the other population. The source/target splits we consider are as follows:

\begin{enumerate}
    \item Source data: White, More than 12 education years. Target data: Non-white, Less than 12 education years. 
    \item Source data: White, Older than 30. Target data: Non-white, younger than 40.
    \item Source data: Younger than 70, More than 12 education years. Target data: Older than 70, less than 12 years of education. 
\end{enumerate}

In Table \ref{table:data-split-stats} we analyze the label and sensitive attribute conditional distributions for the above data splits. For the random split (A), the training and test label and conditional sensitive attribute distributions are identical, which is to be expected. For the three custom splits we observe all three distributions: $P(Y), P(A | Y=0), P(A | Y=1)$ differ between training and test. We also note the label distribution becomes more skewed towards $Y = 0$. 

\subsection{Fairness Metrics} 
\label{sec:met}

There exist a multitude of criteria developed for evaluating algorithmic fairness \cite{mehrabi2021survey}. 
In the context of datasets presenting a privileged and unprivileged group, these metrics rely on ensuring predictive parity between the two groups under different constraints. 
The most common fairness metric employed is demographic parity (DP) $P(\hat Y = 1 | A = 0) = P(\hat Y = 1 | A = 1)$, which is optimized when predicted label probability is identical across the two groups. 
However, DP only ensures similar representation between the two groups, while ignoring actual label distribution. 
Equal opportunity (EO) \cite{hardt2016equality} conditions the fairness value on the true label $Y$, and is optimized when $P(\hat Y = 1 | A = 0, Y = 1) = P(\hat Y = 1 | A = 1, Y = 1)$. EO is preferred when the label distribution is different across privilege classes, i.e. $P(Y | A = 0) \neq P(Y | A = 1)$. A more constrained fairness metric is averaged odds (AO), which is minimized when outcomes are the same conditioned on both labels and sensitive attributes i.e. $P(\hat Y | A = 0, Y = y) = P(\hat Y | A = 1, Y = y), y \in \{0, 1\}$. 
EO is a special case of AO, for the case where $y = 1$. Following fairness literature, we will report the left hand side and right hand side difference $\Delta$ for each of the above measures. Under this format, $\Delta$ values close to $0$ will signify the model maintains fairness, while values close to $1$ signify a lack of fairness. 
Tuning a model to optimize fairness may incur accuracy trade offs \cite{madras2018learning,kleinberg2016inherent,wick2019unlocking}. For example, a classifier which predicts every element to be part of the same group e.g. $P(\hat Y = 0) = 1$ will obtain $\Delta EO = \Delta EO = \Delta AO = 0$, without providing informative predictions. Our approach has the advantage that the regularizers of the three employed losses $\mathcal{L}_{CE}, \mathcal{L}_{fair}, \mathcal{L}_{swd}$ can be tuned in accordance with the importance of accuracy against fairness for a specific task.

\subsection{Other methods}

We evaluate our work against four popular fairness preserving algorithms part of the AIF360 \cite{aif360-oct-2018} project: Meta-Algorithm for Fair Classification (MC) \cite{10.1145/3287560.3287586}, Adversarial Debiasing (AD) \cite{10.1145/3278721.3278779}, Reject Option Classification \cite{6413831} and Exponentiated Gradient Reduction \cite{pmlr-v80-agarwal18a}. We additionally report as baseline (Base) the version of our algorithm where we only minimize $\mathcal{L}_{bce}$, without optimizing fairness or minimizing distributional distance. This corresponds to the performance of a naive source trained classifier. 

\subsection{Results}

We present results for three challenging data splits for each of the considered datasets. We report balanced accuracy (Acc.), demographic parity ($\Delta DP$), equalized odds ($\Delta EO$) and averaged opportunity ($\Delta AO$). Desirable accuracy values are close to $1$, while desirable fairness metric values should be close to $0$. Results are averaged over $7$ runs. Unless otherwise specified, we use \textit{sex} as the sensitive attribute $A$, as it is common for all datasets. 

\begin{table*}[!ht]
    \centering
    \scalebox{.85}{
        \setlength\tabcolsep{2pt}
        \begin{tabular}{|l|c|c|c|c||c|c|c|c||c|c|c|c|}
        \hline
        \multicolumn{1}{|c|}{Alg.} & \multicolumn{4}{c|}{Race, Education} & \multicolumn{4}{c|}{Race, Age} & \multicolumn{4}{c|}{Age, Education} \\
        \cline{2-13}
        &Acc. &$\Delta DP$ &$\Delta EO$ &$\Delta AO$ &Acc. &$\Delta DP$ &$\Delta EO$ &$\Delta AO$ &Acc. &$\Delta DP$ &$\Delta EO$ &$\Delta AO$ \\
        \hline
        Base & 0.63 & 0.34 & 0.53 & 0.42 & 0.60 & 0.24 & 0.25 & 0.24 & 0.59 & 0.90 & 0.92 & 0.91 \\
        MC & 0.68 & 0.28 & 0.32 & 0.28 & 0.63 & 0.05 & 0.26 & 0.15 & 0.63 & 1.00 & 1.00 & 1.00 \\
        AD & 0.63 & 0.21 & 0.33 & 0.25 & 0.60 & 0.25 & 0.25 & 0.25 & 0.51 & 0.16 & 0.15 & 0.16 \\
        ROC & 0.59 & 0.34 & 0.25 & 0.31 & 0.62 & 0.02 & 0.20 & 0.11 & 0.50 & 0.00 & 0.00 & 0.00 \\
        EGR & 0.62 & 0.06 & 0.16 & 0.10 & 0.59 & 0.02 & 0.19 & 0.11 & 0.56 & 0.43 & 0.40 & 0.42 \\
        \hline
        Ours & 0.62 & 0.01 & 0.05 & 0.01 & 0.62 & 0.00 & 0.19 & 0.10 & 0.52 & 0.01 & 0.06 & 0.03 \\
        \hline
        \end{tabular}
    }
    \caption{Performance results for the three splits of the Adult dataset}
    \label{table:Adult-results}
\end{table*}

\textbf{Adult dataset} We report results on the Adult dataset in Table \ref{table:Adult-results}. For each of the custom splits, we report accuracy and fairness metrics. On the first split, MC obtains the highest accuracy of $0.68$, and AD obtains accuracy of $0.63$. Both are higher than our method, however these methods do not maintain fairness, as can be seen by the large $\Delta DP, \Delta EO, \Delta AO$ values. On the remaining tasks our method is able to maintain fairness after adaptation, while being competitive in terms of accuracy. This shows that previous fairness preserving classifiers struggle with domain shift between the source and target, while our method is positioned to overcome these issues.


\begin{table*}[!ht]
    \centering
    \scalebox{.85}{
        \setlength\tabcolsep{2pt}
        \begin{tabular}{|l|c|c|c|c||c|c|c|c||c|c|c|c|}
        \hline
        \multicolumn{1}{|c|}{Alg.} & \multicolumn{4}{c|}{Age, Priors} & \multicolumn{4}{c|}{Race, Age, Priors} & \multicolumn{4}{c|}{Race, Age, Prrs., Chrg.} \\
        \cline{2-13}
        &Acc. &$\Delta DP$ &$\Delta EO$ &$\Delta AO$ &Acc. &$\Delta DP$ &$\Delta EO$ &$\Delta AO$ &Acc. &$\Delta DP$ &$\Delta EO$ &$\Delta AO$ \\
        \hline
        Base & 0.54 & 0.29 & 0.27 & 0.28 & 0.49 & 0.33 & 0.56 & 0.43 & 0.57 & 1.00 & 1.00 & 1.00 \\
        MC & 0.58 & 0.33 & 0.36 & 0.33 & 0.50 & 0.00 & 0.00 & 0.00 & 0.50 & 0.19 & 0.19 & 0.19 \\
        AD & 0.52 & 0.62 & 0.73 & 0.66 & 0.47 & 0.70 & 0.72 & 0.70 & 0.44 & 0.87 & 0.88 & 0.87 \\
        ROC & 0.53 & 0.28 & 0.09 & 0.21 & 0.50 & 0.00 & 0.00 & 0.00 & 0.55 & 0.47 & 0.38 & 0.44 \\
        EGR & 0.49 & 0.05 & 0.10 & 0.06 & 0.53 & 0.27 & 0.34 & 0.26 & 0.50 & 0.00 & 0.00 & 0.00 \\
        \hline
        Ours & 0.53 & 0.00 & 0.05 & 0.02 & 0.65 & 0.15 & 0.17 & 0.19 & 0.50 & 1.00 & 1.00 & 1.00 \\
        \hline
        \end{tabular}
    }
    \caption{Performance results for the three splits of the  COMPAS dataset}
    \label{table:compas-results}
\end{table*}

\textbf{COMPAS dataset} The COMPAS dataset results are reported in Table \ref{table:compas-results}. On the first data split the MC method achieves best accuracy, however none of the methods we compare against besides EGR are able to preserve fairness. We are able to obtain higher accuracy than EGR while also obtaining improved fairness scores. 
On the second data split our method is able to achieve the highest accuracy and also the lowest fairness scores amongst the considered methods. EGR, AD and Base are not able to maintain fairness, while MC and ROC provide degenerate results. 
On the last data split the best results in terms of fairness correspond to MC, and the best accuracy out of the fairness preserving algorithms is provided by the ROC. Other approaches, ours included, provide degenerate solutions. This split has the smallest size amongst all considered splits. Our method relies on minimizing a probabilistic distributional metric for adaptation, task which becomes harder in extremely low data regimes without additional assumptions or regularizers. 

\begin{table*}[!ht]
    \centering
    \scalebox{.85}{
        \setlength\tabcolsep{2pt}
        \begin{tabular}{|l|c|c|c|c||c|c|c|c||c|c|c|c|}
        \hline
        \multicolumn{1}{|c|}{Alg.} & \multicolumn{4}{c|}{Employment} & \multicolumn{4}{c|}{Credit hist., Empl.} & \multicolumn{4}{c|}{Credit hist., Empl.} \\
        \cline{2-13}
        &Acc. &$\Delta DP$ &$\Delta EO$ &$\Delta AO$ &Acc. &$\Delta DP$ &$\Delta EO$ &$\Delta AO$ &Acc. &$\Delta DP$ &$\Delta EO$ &$\Delta AO$ \\
        \hline
        Base & 0.67 & 0.09 & 0.05 & 0.07 & 0.58 & 0.07 & 0.10 & 0.06 & 0.56 & 0.35 & 0.35 & 0.32 \\
        MC & 0.67 & 0.06 & 0.12 & 0.03 & 0.56 & 0.15 & 0.34 & 0.22 & 0.55 & 0.30 & 0.34 & 0.30 \\
        AD & 0.52 & 0.53 & 0.58 & 0.55 & 0.53 & 0.40 & 0.56 & 0.46 & 0.52 & 0.44 & 0.52 & 0.46 \\
        ROC & 0.50 & 0.00 & 0.00 & 0.00 & 0.50 & 0.00 & 0.00 & 0.00 & 0.62 & 0.13 & 0.09 & 0.11 \\
        EGR & 0.57 & 0.22 & 0.36 & 0.27 & 0.50 & 0.43 & 0.44 & 0.43 & 0.50 & 0.01 & 0.00 & 0.00 \\
        \hline
        Ours & 0.62 & 0.01 & 0.05 & 0.02 & 0.58 & 0.02 & 0.01 & 0.01 & 0.55 & 0.01 & 0.02 & 0.01 \\
        \hline
        \end{tabular}
    }
    \caption{Performance results for the three splits of the German dataset}
    \label{table:German-results}
\end{table*}

\begin{figure*}[!htb]
    \centering
    \includegraphics[width=.4\textwidth]{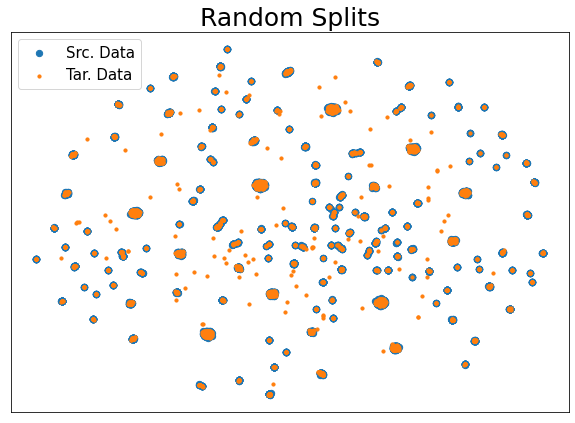}
    \includegraphics[width=.4\textwidth]{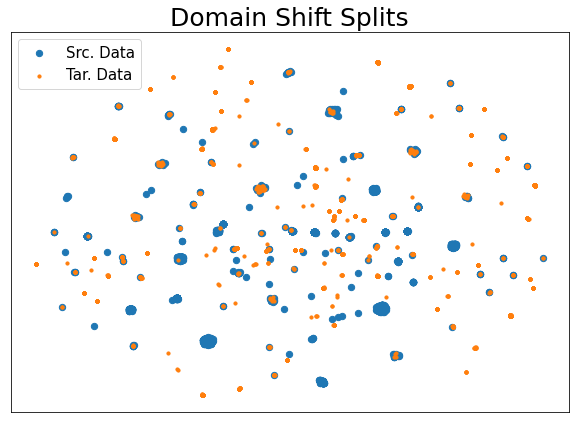}
    \caption{UMAP embeddings of the source and target feature spaces for random and custom splits of the Adult dataset}
    \label{figure:umap-embedding}
\end{figure*}

\textbf{German dataset} We report results on the German dataset in Table \ref{table:German-results}. On the first data split the best accuracy is obtained by MC. Our method obtains second best accuracy, with improved fairness performance. On the second data split our method obtains highest accuracy and best fairness performance across all fairness metrics. 
On the last data split we obtain the best fairness performance. 
EGR does not provide informative predictions, MC,AD do not maintain fairness, and ROC improves accuracy for a fairness trade-off.


\subsection{Ablative Experiments}

We visualize source-target domain shift by generating $2D$ embeddings of the source and target feature spaces corresponding to different splits of the data. For this task, we employ the UMAP \cite{mcinnes2020umap} visualization tool. In Figure \ref{figure:umap-embedding} we compare the source and target features resulting from a random split of the Adult dataset, to our first custom split i.e. by race and education. In case of randomly splitting the dataset, we observe that source and target samples visually share a similar feature space. However, in the case of the custom split, there is significant discrepancy between the two distributions, which affects model generalization. This observation is in line with our numerical results, suggesting that when domain shift is present, model generalization and fairness transfer become more difficult.

\section{Conclusions and Future Work}

Fairness preserving methods have overlooked the problem of domain shift when deploying a source trained model to a target domain. Our first contribution is providing different data splits for popular datasets employed in fairness tasks which present significant domain shift between the source and target. We show that as the distribution of data changes between the two domains, popular fairness-preserving algorithms cannot match the performance they observe on random data splits, where the source and target features are sampled from the same distribution. This observation demonstrates that model fairness is not naturally preserved under domain shift. Second, we present a novel algorithm that addresses domain shift when a fair outcome is of concern by combining the following techniques: (1) fair model training via adversarially generating a sensitive attribute-independent latent feature space, (2) producing a shared latent feature space for the source and target domains by minimizing an appropriate probability distance metric between the source and target embedding distributions. Through empirical evaluation, we show that combining these two ideas ensures a fair model trained on a source domain can generalize on a target domain while maintaining fairness preserving properties even in the presence of domain shift. Future   extensions include investigating scenarios where in addition to maintaining fairness, data may only be accessed sequentially, necessitating source-free model updating~\cite{stan2021unsupervised}.

\clearpage

\bibliographystyle{named}
\bibliography{ijcai23}

\begin{thebibliography}{}

\bibitem[\protect\citeauthoryear{Agarwal \bgroup \em et al.\egroup
  }{2018}]{pmlr-v80-agarwal18a}
Alekh Agarwal, Alina Beygelzimer, Miroslav Dudik, John Langford, and Hanna
  Wallach.
\newblock A reductions approach to fair classification.
\newblock In Jennifer Dy and Andreas Krause, editors, {\em Proceedings of the
  35th International Conference on Machine Learning}, volume~80 of {\em
  Proceedings of Machine Learning Research}, pages 60--69. PMLR, 10--15 Jul
  2018.

\bibitem[\protect\citeauthoryear{Bellamy \bgroup \em et al.\egroup
  }{2018}]{aif360-oct-2018}
Rachel K.~E. Bellamy, Kuntal Dey, Michael Hind, Samuel~C. Hoffman, Stephanie
  Houde, Kalapriya Kannan, Pranay Lohia, Jacquelyn Martino, Sameep Mehta,
  Aleksandra Mojsilovic, Seema Nagar, Karthikeyan~Natesan Ramamurthy, John
  Richards, Diptikalyan Saha, Prasanna Sattigeri, Moninder Singh, Kush~R.
  Varshney, and Yunfeng Zhang.
\newblock {AI Fairness} 360: An extensible toolkit for detecting,
  understanding, and mitigating unwanted algorithmic bias, October 2018.

\bibitem[\protect\citeauthoryear{Beutel \bgroup \em et al.\egroup
  }{2017}]{beutel2017data}
Alex Beutel, Jilin Chen, Zhe Zhao, and Ed~H Chi.
\newblock Data decisions and theoretical implications when adversarially
  learning fair representations.
\newblock {\em arXiv preprint arXiv:1707.00075}, 2017.

\bibitem[\protect\citeauthoryear{Bhushan~Damodaran \bgroup \em et al.\egroup
  }{2018}]{damodaran2018deepjdot}
Bharath Bhushan~Damodaran, Benjamin Kellenberger, R{\'e}mi Flamary, Devis Tuia,
  and Nicolas Courty.
\newblock Deepjdot: Deep joint distribution optimal transport for unsupervised
  domain adaptation.
\newblock In {\em Proceedings of the European Conference on Computer Vision
  (ECCV)}, pages 447--463, 2018.

\bibitem[\protect\citeauthoryear{Buolamwini and
  Gebru}{2018}]{buolamwini2018gender}
Joy Buolamwini and Timnit Gebru.
\newblock Gender shades: Intersectional accuracy disparities in commercial
  gender classification.
\newblock In {\em Conference on fairness, accountability and transparency},
  pages 77--91. PMLR, 2018.

\bibitem[\protect\citeauthoryear{Caliskan \bgroup \em et al.\egroup
  }{2017}]{caliskan2017semantics}
Aylin Caliskan, Joanna~J Bryson, and Arvind Narayanan.
\newblock Semantics derived automatically from language corpora contain
  human-like biases.
\newblock {\em Science}, 356(6334):183--186, 2017.

\bibitem[\protect\citeauthoryear{Celis \bgroup \em et al.\egroup
  }{2019a}]{celis2019classification}
L~Elisa Celis, Lingxiao Huang, Vijay Keswani, and Nisheeth~K Vishnoi.
\newblock Classification with fairness constraints: A meta-algorithm with
  provable guarantees.
\newblock In {\em Proceedings of the conference on fairness, accountability,
  and transparency}, pages 319--328, 2019.

\bibitem[\protect\citeauthoryear{Celis \bgroup \em et al.\egroup
  }{2019b}]{10.1145/3287560.3287586}
L.~Elisa Celis, Lingxiao Huang, Vijay Keswani, and Nisheeth~K. Vishnoi.
\newblock Classification with fairness constraints: A meta-algorithm with
  provable guarantees.
\newblock In {\em Proceedings of the Conference on Fairness, Accountability,
  and Transparency}, FAT* '19, page 319–328, New York, NY, USA, 2019.
  Association for Computing Machinery.

\bibitem[\protect\citeauthoryear{Chouldechova and
  Roth}{2018}]{chouldechova2018frontiers}
Alexandra Chouldechova and Aaron Roth.
\newblock The frontiers of fairness in machine learning.
\newblock {\em arXiv preprint arXiv:1810.08810}, 2018.

\bibitem[\protect\citeauthoryear{Coston \bgroup \em et al.\egroup
  }{2019}]{coston2019fair}
Amanda Coston, Karthikeyan~Natesan Ramamurthy, Dennis Wei, Kush~R Varshney,
  Skyler Speakman, Zairah Mustahsan, and Supriyo Chakraborty.
\newblock Fair transfer learning with missing protected attributes.
\newblock In {\em Proceedings of the 2019 AAAI/ACM Conference on AI, Ethics,
  and Society}, pages 91--98, 2019.

\bibitem[\protect\citeauthoryear{Courty \bgroup \em et al.\egroup
  }{2016}]{courty2017optimal}
Nicolas Courty, R{\'e}mi Flamary, Devis Tuia, and Alain Rakotomamonjy.
\newblock Optimal transport for domain adaptation.
\newblock {\em IEEE Transactions on Pattern Analysis and Machine Intelligence},
  39(9):1853--1865, 2016.

\bibitem[\protect\citeauthoryear{Daum{\'e}~III}{2009}]{daume2009frustratingly}
Hal Daum{\'e}~III.
\newblock Frustratingly easy domain adaptation.
\newblock {\em arXiv preprint arXiv:0907.1815}, 2009.

\bibitem[\protect\citeauthoryear{Du \bgroup \em et al.\egroup
  }{2021}]{du2021fairness}
Mengnan Du, Subhabrata Mukherjee, Guanchu Wang, Ruixiang Tang, Ahmed Awadallah,
  and Xia Hu.
\newblock Fairness via representation neutralization.
\newblock {\em Advances in Neural Information Processing Systems}, 34, 2021.

\bibitem[\protect\citeauthoryear{Dua and Graff}{2017}]{Dua:2019}
Dheeru Dua and Casey Graff.
\newblock {UCI} machine learning repository, 2017.

\bibitem[\protect\citeauthoryear{Feldman \bgroup \em et al.\egroup
  }{2015}]{feldman2015certifying}
Michael Feldman, Sorelle~A Friedler, John Moeller, Carlos Scheidegger, and
  Suresh Venkatasubramanian.
\newblock Certifying and removing disparate impact.
\newblock In {\em proceedings of the 21th ACM SIGKDD international conference
  on knowledge discovery and data mining}, pages 259--268, 2015.

\bibitem[\protect\citeauthoryear{Hardt \bgroup \em et al.\egroup
  }{2016}]{hardt2016equality}
Moritz Hardt, Eric Price, and Nati Srebro.
\newblock Equality of opportunity in supervised learning.
\newblock {\em Advances in neural information processing systems}, 29, 2016.

\bibitem[\protect\citeauthoryear{Hu \bgroup \em et al.\egroup
  }{2019}]{hu2019distributed}
Hui Hu, Yijun Liu, Zhen Wang, and Chao Lan.
\newblock A distributed fair machine learning framework with private
  demographic data protection.
\newblock In {\em 2019 IEEE International Conference on Data Mining (ICDM)},
  pages 1102--1107. IEEE, 2019.

\bibitem[\protect\citeauthoryear{Jiang \bgroup \em et al.\egroup
  }{2020}]{jiang2020wasserstein}
Ray Jiang, Aldo Pacchiano, Tom Stepleton, Heinrich Jiang, and Silvia Chiappa.
\newblock Wasserstein fair classification.
\newblock In {\em Uncertainty in Artificial Intelligence}, pages 862--872.
  PMLR, 2020.

\bibitem[\protect\citeauthoryear{Kamiran \bgroup \em et al.\egroup
  }{2012}]{6413831}
Faisal Kamiran, Asim Karim, and Xiangliang Zhang.
\newblock Decision theory for discrimination-aware classification.
\newblock In {\em 2012 IEEE 12th International Conference on Data Mining},
  pages 924--929, 2012.

\bibitem[\protect\citeauthoryear{Kleinberg \bgroup \em et al.\egroup
  }{2016}]{kleinberg2016inherent}
Jon Kleinberg, Sendhil Mullainathan, and Manish Raghavan.
\newblock Inherent trade-offs in the fair determination of risk scores.
\newblock {\em arXiv preprint arXiv:1609.05807}, 2016.

\bibitem[\protect\citeauthoryear{Lee \bgroup \em et al.\egroup
  }{2019}]{Lee_2019_CVPR}
Chen-Yu Lee, Tanmay Batra, Mohammad~Haris Baig, and Daniel Ulbricht.
\newblock Sliced wasserstein discrepancy for unsupervised domain adaptation.
\newblock In {\em Proceedings of the IEEE/CVF Conference on Computer Vision and
  Pattern Recognition (CVPR)}, June 2019.

\bibitem[\protect\citeauthoryear{Long \bgroup \em et al.\egroup
  }{2015}]{long2015learning}
Mingsheng Long, Yue Cao, Jianmin Wang, and Michael Jordan.
\newblock Learning transferable features with deep adaptation networks.
\newblock In {\em Proceedings of International Conference on Machine Learning},
  pages 97--105, 2015.

\bibitem[\protect\citeauthoryear{Long \bgroup \em et al.\egroup
  }{2017}]{long2017deep}
Mingsheng Long, Han Zhu, Jianmin Wang, and Michael~I Jordan.
\newblock Deep transfer learning with joint adaptation networks.
\newblock In {\em Proceedings of the 34th International Conference on Machine
  Learning-Volume 70}, pages 2208--2217. JMLR. org, 2017.

\bibitem[\protect\citeauthoryear{Madras \bgroup \em et al.\egroup
  }{2018}]{madras2018learning}
David Madras, Elliot Creager, Toniann Pitassi, and Richard Zemel.
\newblock Learning adversarially fair and transferable representations.
\newblock In {\em International Conference on Machine Learning}, pages
  3384--3393. PMLR, 2018.

\bibitem[\protect\citeauthoryear{McInnes \bgroup \em et al.\egroup
  }{2020}]{mcinnes2020umap}
Leland McInnes, John Healy, and James Melville.
\newblock Umap: Uniform manifold approximation and projection for dimension
  reduction, 2020.

\bibitem[\protect\citeauthoryear{Mehrabi \bgroup \em et al.\egroup
  }{2021}]{mehrabi2021survey}
Ninareh Mehrabi, Fred Morstatter, Nripsuta Saxena, Kristina Lerman, and Aram
  Galstyan.
\newblock A survey on bias and fairness in machine learning.
\newblock {\em ACM Computing Surveys (CSUR)}, 54(6):1--35, 2021.

\bibitem[\protect\citeauthoryear{Paszke \bgroup \em et al.\egroup
  }{2019}]{NEURIPS2019_9015}
Adam Paszke, Sam Gross, Francisco Massa, Adam Lerer, James Bradbury, Gregory
  Chanan, Trevor Killeen, Zeming Lin, Natalia Gimelshein, Luca Antiga, Alban
  Desmaison, Andreas Kopf, Edward Yang, Zachary DeVito, Martin Raison, Alykhan
  Tejani, Sasank Chilamkurthy, Benoit Steiner, Lu~Fang, Junjie Bai, and Soumith
  Chintala.
\newblock Pytorch: An imperative style, high-performance deep learning library.
\newblock In H.~Wallach, H.~Larochelle, A.~Beygelzimer, F.~d\textquotesingle
  Alch\'{e}-Buc, E.~Fox, and R.~Garnett, editors, {\em Advances in Neural
  Information Processing Systems 32}, pages 8024--8035. Curran Associates,
  Inc., 2019.

\bibitem[\protect\citeauthoryear{Pooch \bgroup \em et al.\egroup
  }{2019}]{pooch2019can}
Eduardo~HP Pooch, Pedro~L Ballester, and Rodrigo~C Barros.
\newblock Can we trust deep learning models diagnosis? the impact of domain
  shift in chest radiograph classification.
\newblock {\em arXiv preprint arXiv:1909.01940}, 2019.

\bibitem[\protect\citeauthoryear{Redko \bgroup \em et al.\egroup
  }{2017}]{10.1007/978-3-319-71246-8_45}
Ievgen Redko, Amaury Habrard, and Marc Sebban.
\newblock Theoretical analysis of domain adaptation with optimal transport.
\newblock In Michelangelo Ceci, Jaakko Hollm{\'e}n, Ljup{\v{c}}o Todorovski,
  Celine Vens, and Sa{\v{s}}o D{\v{z}}eroski, editors, {\em Machine Learning
  and Knowledge Discovery in Databases}, pages 737--753, Cham, 2017. Springer
  International Publishing.

\bibitem[\protect\citeauthoryear{Rostami and Galstyan}{2023}]{rostami2023}
Mohammad Rostami and Aram Galstyan.
\newblock Overcoming concept shift in domain-aware settings through
  consolidated internal distributions.
\newblock In {\em Proceedings of the AAAI conference on artificial
  intelligence}, 2023.

\bibitem[\protect\citeauthoryear{Rostami \bgroup \em et al.\egroup
  }{2018}]{rostami2018crowdsourcing}
Mohammad Rostami, David Huber, and Tsai-Ching Lu.
\newblock A crowdsourcing triage algorithm for geopolitical event forecasting.
\newblock In {\em Proceedings of the 12th ACM Conference on Recommender
  Systems}, pages 377--381, 2018.

\bibitem[\protect\citeauthoryear{Rostami \bgroup \em et al.\egroup
  }{2020}]{rostami2020using}
Mohammad Rostami, David Isele, and Eric Eaton.
\newblock Using task descriptions in lifelong machine learning for improved
  performance and zero-shot transfer.
\newblock {\em Journal of Artificial Intelligence Research}, 67:673--704, 2020.

\bibitem[\protect\citeauthoryear{Rostami}{2021}]{rostami2021lifelong}
Mohammad Rostami.
\newblock Lifelong domain adaptation via consolidated internal distribution.
\newblock {\em Advances in neural information processing systems},
  34:11172--11183, 2021.

\bibitem[\protect\citeauthoryear{Schumann \bgroup \em et al.\egroup
  }{2019}]{schumann2019transfer}
Candice Schumann, Xuezhi Wang, Alex Beutel, Jilin Chen, Hai Qian, and Ed~H Chi.
\newblock Transfer of machine learning fairness across domains.
\newblock 2019.

\bibitem[\protect\citeauthoryear{Singh \bgroup \em et al.\egroup
  }{2021}]{singh2021fairness}
Harvineet Singh, Rina Singh, Vishwali Mhasawade, and Rumi Chunara.
\newblock Fairness violations and mitigation under covariate shift.
\newblock In {\em Proceedings of the 2021 ACM Conference on Fairness,
  Accountability, and Transparency}, pages 3--13, 2021.

\bibitem[\protect\citeauthoryear{Stan and Rostami}{2021}]{stan2021unsupervised}
Serban Stan and Mohammad Rostami.
\newblock Unsupervised model adaptation for continual semantic segmentation.
\newblock In {\em Proceedings of the AAAI conference on artificial
  intelligence}, volume~35, pages 2593--2601, 2021.

\bibitem[\protect\citeauthoryear{Stan and Rostami}{2022a}]{stan2022domain}
Serban Stan and Mohammad Rostami.
\newblock Domain adaptation for the segmentation of confidential medical
  images.
\newblock In {\em British Machine Vision Conference}, 2022.

\bibitem[\protect\citeauthoryear{Stan and Rostami}{2022b}]{stansecure}
Serban Stan and Mohammad Rostami.
\newblock Secure domain adaptation with multiple sources.
\newblock {\em Transactions on Machine Learning Research}, 2022.

\bibitem[\protect\citeauthoryear{Sun and Saenko}{2016}]{sun2016deep}
Baochen Sun and Kate Saenko.
\newblock Deep coral: Correlation alignment for deep domain adaptation.
\newblock In {\em European conference on computer vision}, pages 443--450.
  Springer, 2016.

\bibitem[\protect\citeauthoryear{Tzeng \bgroup \em et al.\egroup
  }{2017}]{tzeng2017adversarial}
Eric Tzeng, Judy Hoffman, Kate Saenko, and Trevor Darrell.
\newblock Adversarial discriminative domain adaptation.
\newblock In {\em Proceedings of the IEEE conference on computer vision and
  pattern recognition}, pages 7167--7176, 2017.

\bibitem[\protect\citeauthoryear{Wick \bgroup \em et al.\egroup
  }{2019}]{wick2019unlocking}
Michael Wick, Jean-Baptiste Tristan, et~al.
\newblock Unlocking fairness: a trade-off revisited.
\newblock {\em Advances in neural information processing systems}, 32, 2019.

\bibitem[\protect\citeauthoryear{Zhang and Long}{2021}]{zhang2021assessing}
Yiliang Zhang and Qi~Long.
\newblock Assessing fairness in the presence of missing data.
\newblock {\em Advances in Neural Information Processing Systems}, 34, 2021.

\bibitem[\protect\citeauthoryear{Zhang \bgroup \em et al.\egroup
  }{2018}]{10.1145/3278721.3278779}
Brian~Hu Zhang, Blake Lemoine, and Margaret Mitchell.
\newblock Mitigating unwanted biases with adversarial learning.
\newblock AIES '18, page 335–340, New York, NY, USA, 2018. Association for
  Computing Machinery.

\end{thebibliography}

\clearpage

\section{Appendix}

\begin{table*}[!ht]
    \centering
    \scalebox{.9}{
        \setlength\tabcolsep{2pt}
        \begin{tabular}{|l|c|c|c|c|c|c|c|c|c|c|c|c|}
        \hline
        \multicolumn{1}{|c|}{Split} & \multicolumn{4}{c|}{Source} & \multicolumn{4}{c|}{Target} \\
        \cline{2-9}
        & Size &Y=0 &A=0$|$Y=0 &A=0$|$Y=1 &Size &Y=0  &A=0$|$Y=0 &A=0$|$Y=1 \\
        \hline
        A & 34120 & 0.76 & 0.39 & 0.15 & 14722 & 0.76 & 0.39 & 0.15 \\
        A1 & 12024 & 0.53 & 0.41 & 0.16 & 5393 & 0.91 & 0.49 & 0.18 \\
        A2 & 29466 & 0.66 & 0.34 & 0.14 & 2219 & 0.97 & 0.48 & 0.30 \\
        A3 & 11887 & 0.52 & 0.42 & 0.16 & 778 & 0.89 & 0.39 & 0.17 \\
        \hline
        C & 3701 & 0.52 & 0.77 & 0.86 & 1577 & 0.54 & 0.76 & 0.84 \\
        C1 & 2886 & 0.58 & 0.74 & 0.82 & 1096 & 0.67 & 0.78 & 0.86 \\
        C2 & 903 & 0.47 & 0.80 & 0.80 & 96 & 0.74 & 0.70 & 0.92 \\
        C3 & 292 & 0.51 & 0.77 & 0.79 & 50 & 0.68 & 0.62 & 0.88 \\
        \hline
        G & 697 & 0.70 & 0.28 & 0.37 & 303 & 0.70 & 0.30 & 0.34 \\
        G1 & 573 & 0.66 & 0.34 & 0.45 & 427 & 0.76 & 0.23 & 0.20 \\
        G2 & 388 & 0.61 & 0.36 & 0.49 & 196 & 0.84 & 0.20 & 0.16 \\
        G3 & 439 & 0.62 & 0.35 & 0.45 & 159 & 0.87 & 0.21 & 0.19 \\
        \hline
        \end{tabular}
    }
    \caption{Data split statistics. A,C,G correspond to the Adult, COMPAS and German dataset respectively. The rows with no number i.e. A,C,G correspond to random data splits. The numbered rows e.g. A1,A2,A3 correspond to statistics for specific splits. The columns represent the probabilities of specific outcomes for specific splits e.g. $P(Y = 0)$. Results when using \textit{sex} as sensitive attribute.}
    \label{table:data-split-stats-full}
\end{table*}

\subsection{Data splits}

The data splits employed in our approach are as follows:

\textbf{Adult Dataset}. We will use age, education and race to generate source and target domains. This can be a natural occurrence in practice, as gathered census information may differ along these axes geographically. For example, urban population is on average more educate than rural population \footnote{https://www.ers.usda.gov/topics/rural-economy-population/employment-education/rural-education/}, and more ethnically diverse \footnote{https://www.ers.usda.gov/data-products/chart-gallery/gallery/chart-detail/?chartId=99538}. Thus, a fair model trained on one of the two populations will need to overcome distribution shift when evaluated on the other population. Besides differences in the feature distributions, we also note the Adult dataset is both imbalanced in terms of outcome, $P(Y = 1) = 0.34$, and sensitive attribute of positive outcome, $P(A = 1 | Y = 1) = 0.85$, i.e. only a fraction of participants are earning more than $50k$/year, and $85\%$ of them are male.  

The source/target splits we consider are as follows:

\begin{enumerate}
    \item Source data: White, More than 12 education years. Target data: Non-white, Less than 12 education years. 
    \item Source data: White, Older than 30. Target data: Non-white, younger than 40.
    \item Source data: Younger than 70, More than 12 education years. Target data: Older than 70, less than 12 years of education. 
\end{enumerate}

In Table \ref{table:data-split-stats-full} we analyze the label and sensitive attribute conditional distributions for the above data splits. For the random split (A), the training and test label and conditional sensitive attribute distributions are identical, which is to be expected. For the three custom splits we observe all three distributions: $P(Y), P(A | Y=0), P(A | Y=1)$ differ between training and test. We also note the label distribution becomes more skewed towards $Y = 0$. 

\textbf{COMPAS Dataset} Compared to the Adult dataset, the COMPAS dataset is balanced in terms of label distribution, however is imbalanced in terms of the conditional distribution of the sensitive attribute. We will split the dataset along age, number of priors, and charge degree, i.e. whether the person committed a felony or misdemeanor. Considered splits are as follows:

\begin{enumerate}
    \item Source data: Younger than 45, Less than 3 prior convictions. Test data: Older than 45, more than 3 prior convictions.
    \item Source data: Younger than 45, White, At least one prior conviction. Target data: Older than 45, Non-white, No prior conviction.
    \item Source data: Younger than 45, White, At least one prior conviction, convicted for a misdemeanor. Target data: Older than 45, Non-white, No prior conviction, convicted for a felony. 
\end{enumerate}

The first split tests whether a young population with limited number of convictions can be leveraged to fairly predict outcomes for an older population with more convictions. The second split introduces racial bias in the sampling process. In the third split we additionally consider the type of felony committed when splitting the dataset. For all splits, the test datasets become more imbalanced compared to the random split. 

\textbf{German Credit Dataset} The dataset is smallest out of the three considered. For splitting we consider credit history and employment history. Similar to the Adult dataset, the label distribution is skewed towards increased risk i.e. $P(Y = 0) = 0.7$, and individuals of low risk are also skewed towards being part of the privileged group i.e. $P(A = 1 | Y =  1) = 0.63$. We consider the following splits:

\begin{enumerate}
    \item Source data: Employed up to 4 years. Test data: Employed long term (4+ years).
    \item Source data: Up to date credit history, Employed less than 4 years. Target data: un-paid credit, Long term employed. 
    \item Source data: Delayed or paid credit, Employed up to 4 years. Target data: Critical account condition, Long term employment. 
\end{enumerate}

Compared to random data splits, the custom splits reduce label and sensitive attribute imbalance in the source domain, and increase these in the target domain.

\subsection{Parameter tuning and implementation}

\subsubsection{Training and model selection} Implementation of our approach is done using the PyTorch \cite{NEURIPS2019_9015} deep learning library. We model our encoder $e_u$ as a one layer neural network with output space $z \in \mathbb{R}^{20}$. Classifiers $g$ and $h$ are also one layer networks with output space $\in \mathbb{R}^2$. We train our model for $45,000$ iterations, where the first $30,000$ iterations only involve source training. For the first $15,000$ we only perform minimization of the binary cross entropy loss $\mathcal{L}_{bce}$. We introduce source fairness training at iteration $15,000$, and train the fair model, i.e. with respect to both $\mathcal{L}_{bce}$ and $\mathcal{L}_{fair}$, for $15,000$ more iterations. In the last $15,000$ iterations we perform adaptation, where we optimize $\mathcal{L}_{bce}, \mathcal{L}_{fair}$ on the source domain, $\mathcal{L}_{fair}$ on the target domain, and $\mathcal{L}_{swd}$ between the source and target embeddings $e_u((x^s, a^s)), e_u((x^t, a^t))$ respectively. We use a learning rate for $\mathcal{L}_{bce}, \mathcal{L}_{fair}$ of $1e-4$, and learning rate for $\mathcal{L}_{swd}$ of $1e-5$. Model selection is done by considering the difference between accuracy on the validation set, and demographic parity on the test set. Given equalized odds and averaged opportunity require access to the underlying labels on the test set we cannot use these metrics for model selection. Additionally, models corresponding to degenerate predictions i.e. test set predicted labels being either all $0$s or all $1$s are not included in result reporting.

\subsection{Additional Results}

\begin{table*}[!ht]
    \centering
    \scalebox{.9}{
        \setlength\tabcolsep{1.5pt}
        \begin{tabular}{|l|c|c|c|c||c|c|c|c||c|c|c|c|}
        \hline
        \multicolumn{1}{|c|}{Alg.} & \multicolumn{4}{c|}{Adult} & \multicolumn{4}{c|}{German} & \multicolumn{4}{c|}{COMPAS} \\
        \cline{2-13}
        &Acc. &$\Delta DP$ &$\Delta EO$ &$\Delta AO$ &Acc. &$\Delta DP$ &$\Delta EO$ &$\Delta AO$ &Acc. &$\Delta DP$ &$\Delta EO$ &$\Delta AO$ \\
        \hline
        Base & 0.74 & 0.35 & 0.30 & 0.28 & 0.64 & 0.03 & 0.16 & 0.05 & 0.68 & 0.22 & 0.27 & 0.18 \\ 
        MC & 0.71 & 0.13 & 0.09 & 0.08 & 0.63 & 0.22 & 0.18 & 0.20 & 0.65 & 0.22 & 0.22 & 0.20 \\ 
        AD & 0.67 & 0.11 & 0.13 & 0.08 & 0.53 & 0.35 & 0.46 & 0.38 & 0.62 & 0.28 & 0.25 & 0.29 \\ 
        ROC & 0.71 & 0.05 & 0.01 & 0.01 & 0.55 & 0.11 & 0.04 & 0.09 & 0.52 & 0.02 & 0.03 & 0.02 \\ 
        EGR & 0.65 & 0.06 & 0.02 & 0.01 & 0.51 & 0.01 & 0.04 & 0.02 & 0.63 & 0.02 & 0.02 & 0.02 \\ 
        \hline
        Ours & 0.70 & 0.00 & 0.07 & 0.08 & 0.64 & 0.00 & 0.05 & 0.01 & 0.65 & 0.00 & 0.02 & 0.03 \\ 
        \hline
        \end{tabular}
    }
    \caption{Results for random data splits.}
    \label{table:random-splits}
\end{table*}

\par We consider results obtained for random splits on the Adult, German and COMPAS datasets, detailed in Table \ref{table:random-splits}. These splits correspond to standard experiments intended to verify whether a fair model will translate fairness benefits to the test data. Under this scenario, the source and target features are sampled from the same distribution, and there is no domain shift. 
We observe the baseline approach is able to obtain highest accuracy across datasets, however it does not provide any fairness guarantees. Out of the fairness preserving methods, our method is able to match or nearly match best accuracy, while providing perfect demographic parity. 
Even though we do not perform model selection based on the other two fairness metrics due to not having access to the target labels, we are able to match best averaged opportunity on the German dataset, and best equalized odds on the COMPAS dataset. 
This set of results shows that even in scenarios where domain shift is not of concern, our algorithm successfully learns a competitive fair model. 

We additionally investigate the impact of the different components of our approach on the numerical results. In Table \ref{table:compas-ablation} we compare performance on the COMPAS dataset for four variants of our algorithm: (1) Base, similar to the main experiments, where no fairness or distributional minimization metric is used, just minimization of $\mathcal{L}_{bce}$ on the source domain samples (2) SWD, only minimizing $\mathcal{L}_{swd}$ (3) Fair, only training with respect to $\mathcal{L}_{fair}$ on the source and target domains (4) Our complete approach using all fairness and adaptation objectives. Note $\mathcal{L}_{bce}$ is still used in cases (2) and (3). On the third data split results for all three variants are inconclusive either in term of accuracy (Base) or fairness. On the first and second splits, utilizing all losses leads to best performance. On the first split, the Fair only model is able to achieve competitive fairness results at the cost of accuracy. The SWD only approach achieves better accuracy but at the cost of fairness. Combining the two losses leads to improved accuracy over the Fair only model, and also improved fairness. Due to $\mathcal{L}_{swd}$ being minimized at the encoder output space, both classifier and fairness head benefit from a shared source-target feature space. On the second split we observe the SWD only model has poorest performance, and the Fair Only and combined model have similar fairness performance, with the combined model obtaining higher accuracy. This signifies the adversarial fair training process can act as a proxy task during training, improving model generalization. 

\begin{table*}[!htb]
    \centering
    \scalebox{.9}{
        \setlength\tabcolsep{1.5pt}
        \begin{tabular}{|l|c|c|c|c||c|c|c|c||c|c|c|c|}
        \hline
        \multicolumn{1}{|c|}{Alg.} & \multicolumn{4}{c|}{Age, Priors} & \multicolumn{4}{c|}{Race, Age, Priors} & \multicolumn{4}{c|}{Race, Age, Prrs., Chrg.} \\
        \cline{2-13}
        &Acc. &$\Delta DP$ &$\Delta EO$ &$\Delta AO$ &Acc. &$\Delta DP$ &$\Delta EO$ &$\Delta AO$ &Acc. &$\Delta DP$ &$\Delta EO$ &$\Delta AO$ \\
        \hline
        Base & 0.68 & 0.22 & 0.27 & 0.18 & 0.54 & 0.29 & 0.27 & 0.28 & 0.49 & 0.33 & 0.56 & 0.43 \\
        SWD & 0.56 & 0.29 & 0.38 & 0.32 & 0.45 & 0.44 & 0.33 & 0.40 & 0.55 & 1.00 & 1.00 & 1.00 \\
        Fair & 0.50 & 0.01 & 0.08 & 0.04 & 0.64 & 0.15 & 0.17 & 0.19 & 0.41 & 1.00 & 1.00 & 1.00 \\
        \hline
        Ours & 0.53 & 0.00 & 0.05 & 0.02 & 0.65 & 0.15 & 0.17 & 0.19 & 0.50 & 1.00 & 1.00 & 1.00 \\
        \hline
        \end{tabular}
    }
    \caption{Results when selectively using a subset of losses on the COMPAS dataset}
    \label{table:compas-ablation}
\end{table*}

\begin{table*}[!htb]
    \centering
    \scalebox{.9}{
        \setlength\tabcolsep{1.5pt}
        \begin{tabular}{|l|c|c|c|c||c|c|c|c||c|c|c|c|}
        \hline
        \multicolumn{1}{|c|}{Alg.} & \multicolumn{4}{c|}{Empl.} & \multicolumn{4}{c|}{Credit hist., Empl.} & \multicolumn{4}{c|}{Credit hist., Empl.} \\
        \cline{2-13}
        &Acc. &$\Delta DP$ &$\Delta EO$ &$\Delta AO$ &Acc. &$\Delta DP$ &$\Delta EO$ &$\Delta AO$ &Acc. &$\Delta DP$ &$\Delta EO$ &$\Delta AO$ \\
        \hline
        Base & 0.63 & 0.46 & 0.33 & 0.40 & 0.61 & 0.19 & 0.15 & 0.16 & 0.58 & 0.23 & 0.25 & 0.19 \\
        MC & 0.59 & 0.32 & 0.29 & 0.30 & 0.67 & 0.35 & 0.18 & 0.27 & 0.61 & 0.12 & 0.18 & 0.11 \\
        AD & 0.51 & 0.54 & 0.61 & 0.55 & 0.50 & 0.53 & 0.52 & 0.54 & 0.52 & 0.50 & 0.60 & 0.53 \\
        ROC & 0.54 & 0.09 & 0.02 & 0.07 & 0.51 & 0.05 & 0.03 & 0.04 & 0.59 & 0.24 & 0.18 & 0.21 \\
        EGR & 0.50 & 0.03 & 0.03 & 0.03 & 0.56 & 0.12 & 0.23 & 0.16 & 0.50 & 0.02 & 0.01 & 0.01 \\
        \hline
        Ours & 0.62 & 0.02 & 0.09 & 0.02 & 0.59 & 0.02 & 0.16 & 0.06 & 0.62 & 0.02 & 0.18 & 0.04 \\
        \hline
        \end{tabular}
    }
    \caption{Results on the German dataset when optimizing fairness metrics with respect to the \textit{age} sensitive attribute}
    \label{table:German-ablation}
\end{table*}

\begin{figure*}[!htb]
    \centering
    \includegraphics[width=.4\textwidth]{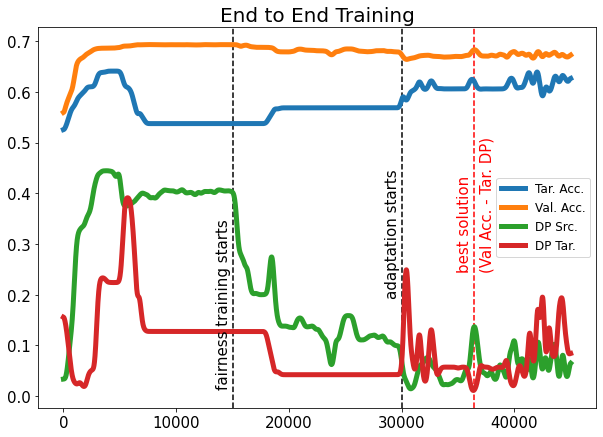}
    \includegraphics[width=.4\textwidth]{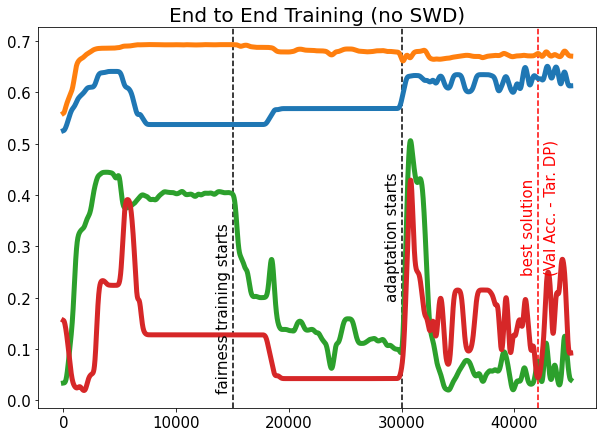}
    \caption{Learning behavior when performing end-to-end training when using both $\mathcal{L}_{fair}$ and $\mathcal{L}_{swd}$ (top) and when only using $\mathcal{L}_{fair}$ (bottom)}
    \label{figure:tar-training}
\end{figure*}

We further investigate the different components present in our algorithm. In Figure \ref{figure:tar-training} we analyze the training and adaptation process with respect to target accuracy, validation accuracy, demographic parity on the source domain, and demographic parity on the target domain. Performance plots are reported for the Adult dataset. We compare two scenarios: running the algorithm when $\mathcal{L}_{swd}$ is not enforced (bottom), and running the algorithm using both fairness and domain alignment (top). For the first $30,000$ iterations we only perform source training, where the first half of iterations is spent optimizing $\mathcal{L}_{bce}$, and the second half is spent jointly optimizing $\mathcal{L}_{bce}$ and the source fairness objective. We note once optimization with respect to $\mathcal{L}_{fair}$ starts, demographic parity decreases until adaptation start, i.e. between iterations $15,000-30,000$. The validation accuracy in this interval also slightly decreases, as improving fairness may affect accuracy performance. During adaptation, i.e. after iteration $30,000$, we observe that in the scenario where we use $\mathcal{L}_{swd}$, the target accuracy increases, while demographic parity on both source and target domains remains relatively unchanged. In the scenario where no optimization of $\mathcal{L}_{swd}$ is performed, there is still improvement with respect to target accuracy, however target demographic parity becomes on average larger. This implies that the distributional alignment loss done at the output of the encoder has beneficial effects both for the classification as well as the fairness objective. 

We further investigate the different components present in our algorithm. In Figure \ref{figure:tar-training} we analyze the training and adaptation process with respect to target accuracy, validation accuracy, demographic parity on the source domain, and demographic parity on the target domain. Performance plots are reported for the Adult dataset. We compare two scenarios: running the algorithm when $\mathcal{L}_{swd}$ is not enforced (bottom), and running the algorithm using both fairness and domain alignment (top). For the first $30,000$ iterations we only perform source training, where the first half of iterations is spent optimizing $\mathcal{L}_{bce}$, and the second half is spent jointly optimizing $\mathcal{L}_{bce}$ and the source fairness objective. We note once optimization with respect to $\mathcal{L}_{fair}$ starts, demographic parity decreases until adaptation start, i.e. between iterations $15,000-30,000$. The validation accuracy in this interval also slightly decreases, as improving fairness may affect accuracy performance. During adaptation, i.e. after iteration $30,000$, we observe that in the scenario where we use $\mathcal{L}_{swd}$, the target accuracy increases, while demographic parity on both source and target domains remains relatively unchanged. In the scenario where no optimization of $\mathcal{L}_{swd}$ is performed, there is still improvement with respect to target accuracy, however target demographic parity becomes on average larger. This implies that the distributional alignment loss done at the output of the encoder has beneficial effects both for the classification as well as the fairness objective. 







\end{document}